\documentclass[conference]{IEEEtran}
\IEEEoverridecommandlockouts
\usepackage{cite}
\usepackage{amsmath,amssymb,amsfonts}
\usepackage{algorithmic}
\usepackage{graphicx}
\usepackage{textcomp}
\usepackage{xcolor}

\usepackage{adjustbox}
\usepackage{url}
\usepackage{hyperref}
\usepackage{subcaption}

\def\BibTeX{{\rm B\kern-.05em{\sc i\kern-.025em b}\kern-.08em
    T\kern-.1667em\lower.7ex\hbox{E}\kern-.125emX}}

\usepackage{tikz}
\usepackage{textcomp}
\usepackage{hyperref}
\usepackage{lipsum}

\newcommand\copyrighttext{%
  \footnotesize \textcopyright 2024 IEEE. Personal use of this material is permitted.
  Permission from IEEE must be obtained for all other uses, in any current or future
  media, including reprinting/republishing this material for advertising or promotional
  purposes, creating new collective works, for resale or redistribution to servers or
  lists, or reuse of any copyrighted component of this work in other works.
  DOI: \href{<https://ieeexplore.ieee.org/document/10645653>}{10.1109/CoG60054.2024.10645653}}
\newcommand\copyrightnotice{%
\begin{tikzpicture}[remember picture,overlay]
\node[anchor=south,yshift=10pt] at (current page.south) {\fbox{\parbox{\dimexpr\textwidth-\fboxsep-\fboxrule\relax}{\copyrighttext}}};
\end{tikzpicture}%
}

\begin{document}

\title{Bayesian Optimization-based Search for Agent Control in Automated Game Testing\\

}

\author{
\IEEEauthorblockN{Carlos Celemin}
\IEEEauthorblockA{\textit{Production R\&D, Tencent} \\
Amsterdam, Netherlands \\
cele@global.tencent.com}

}

\maketitle
\copyrightnotice
\IEEEoverridecommandlockouts

\IEEEpubidadjcol

\begin{abstract}
This work introduces an automated testing approach that employs agents controlling game characters to detect potential bugs within a game level. 
Harnessing the power of Bayesian Optimization (BO) to execute sample-efficient search, the method determines the next sampling point by analyzing the data collected so far and calculates the data point that will maximize information acquisition.
To support the BO process, we introduce a game testing-specific model built on top of a grid map, that features the smoothness and uncertainty estimation required by BO, however and most importantly, it does not suffer the scalability issues that traditional models carry. 
The experiments demonstrate that the approach significantly improves map coverage capabilities in both time efficiency and exploration distribution. 
\end{abstract}


\section{Introduction}
 
There is a spectrum of issues that can be encountered in a game, ranging from the low-level of abstraction, e.g., the related to collisions detection, game mechanics, performance, crash states, all the way to the high-level end problems like game balance, or player experience \cite{bergdahl2020augmenting,zheng2019wuji}.
The massive possible combinations of sequences of game states that could happen in a game scene makes it prohibitively expensive to be able to perform the full game testing process based on human testers.

Some tests can be automated using bots that control the characters that carry out actions in the game.
This has been implemented in most of the cases by engineers who encode manually the behaviors in scripts.
More recently, with the fast progress of Machine Learning (ML), certain techniques have been adopted to develop such behaviors, mainly Reinforcement Learning (RL) and Imitation Learning (IL). 


The contribution of this work aims to improve the efficiency, flexibility, portability, scalability, and robustness in automated game testing systems.
Instead of developing an end-to-end agent that is in charge of exploring the game level, in this work, we propose to implement systems that uncouple decisions with hierarchical modules, i.e., a high-level module makes a decision that conditions the decision of the low-level module.

This short paper is focused on the high-level module of the proposed system, which is a search manager based on Bayesian Optimization (BO), intended to find the most informative map locations an agent should explore given the information acquired so far.
A low-level module based on a given ML behavior takes those locations and tries to reach them, however only the high-level module is in the scope of this work.
We introduce a game testing-specific non-parametric model that makes it feasible to leverage the search capabilities of BO, but without the scalability limitations of the existing models that such optimization uses.

The results of the experimental procedure show that the presented approach features several properties that can improve the efficiency of a game testing process, maximizing the probability of finding bugs, and reducing the cost of exploration.

\begin{figure}
    \centering
    \includegraphics[width=0.95\linewidth]{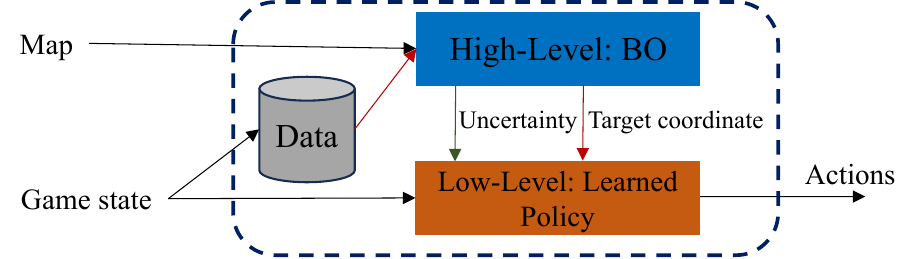}
    \caption{Hierarchical testing system. The high-level model is based on Bayesian Optimization and computes where in the level and how much to explore. The low-level is a learned NN policy that computes the character actions (player level actions). }
    \label{fig:diagram}
\end{figure}

\section{Related Work}
Recently, many works have applied ML to obtain agent behaviors which can be used for either automated testing or implementing NPCs.
RL autonomously learns behaviors by trial and error, and it has been widely applied in video games to obtain agent behaviors \cite{shao2019survey}.
Nevertheless, it has been specifically used for obtaining bots for game testing \cite{bergdahl2020augmenting,shih2020successfully,zheng2019wuji,tufano2022using,gordillo2021improving}.
In \cite{gordillo2021improving}, they use the emergent exploratory behaviors of the learning process itself more than the resulting policy, exploration is encouraged with a curiosity driven learning approach.
Unfortunately, RL processes are computationally and temporally demanding, and require iterating the full process many times to engineer hyperparameters including the objective functions.
All that costly process should be repeated every time important changes are introduced in the levels.

On the other hand, Imitation Learning (IL) techniques train agents from data demonstrated by experts executing the corresponding task, i.e., agents imitating the experts.
It has also been applied to both NPCs and testing agents \cite{pelling2019two,sestini2023towards,gudmundsson2018human,zhao2020winning,huang2020efficient}.
This learning approach is more direct and less expensive for generating desired behaviors, as long as the demonstration data is available.

Although for testing it can be useful  to develop agents that can play well (i.e., agents that try to win or get a high score in a game), such learned behaviours might not be the best suited for the task of finding bugs within a game.
Moreover, depending on the stage of the game development, theoretically sound approaches can suffer robustness issues in practice, for instance, the curiosity driven RL-based agents intended for exploratory behaviors may not work as expected and break the learning process when deployed in buggy environments \cite{gordillo2021improving}.
In our preliminary tests, we have observed that problems in a game level can introduce errors in either the observations or especially the state transitions, introducing inconsistencies that violate the assumptions of the Markov Decision Processes.
Such problems impact the decisions of a trained agent only during the time frames in which the problems are present, however, during training time the inconsistencies of a single frame can impact the performance of the entire learning process.

This work adopts the use of already trained agents (trained in controlled environments), rather than leveraging the emergent behaviors of the RL training process. 
Hence, the decisions of where and how to explore are made by a high-level system, which not only encourages exploration, but does it in an efficient way, by optimizing information acquisition using BO.

\section{Background}

BO is a technique used for finding the maximum (or minimum) of an objective function that is expensive to evaluate, its structure is unknown, and derivative information is not readily available. 
The technique operates by constructing a probabilistic model of the objective function called surrogate model, and then utilizing it to determine where to sample next in the domain of the function. 
This is done by defining an acquisition function that represents the utility of sampling at each point in the search space.
Typically, the most common models used for the surrogate function are Gaussian Processes (GP) and Tree Parzen Estimators (TPE).
Through iterative evaluation of promising points suggested by the acquisition function, BO effectively balances the exploration-exploitation trade-off, searching new areas of the solution space while also exploiting regions known to obtain high returns. 

However, BO has limitations regarding the amount of data the surrogate model can handle, which makes it difficult to scale its applicability to problems that operate with large datasets, as in the case of level exploration in game testing, hindering the use of many existing BO packages for this application.
Hence, this paper proposes a model that can bypass this limitation.

\section{Method}
As previously mentioned, the proposed approach for game level exploration is based on a hierarchical decision-making system (Figure \ref{fig:diagram}).
It features a high-level module in charge of deciding where the agent goes and how much it explores.
Such information along with game states are used by a low-level controller module to compute the sequence of actions required to reach the desired area.
In this case, the low-level behaviors are based on Neural Networks (NN) trained with either IL or RL.
Thus, if the game mechanics are changed, only the low-level module needs to be retrained in the new conditions, but the high-level module remains useful as it is.
Furthermore, if a test aims to search for a different  kind of bug, only the high-level module, which estimates the most promising areas for exploration, need to be adapted. 
The low-level NN behavior can still be reused.
This modular approach improves the portability and flexibility of the testing system, while reducing the training time of the agents that need to achieve only one of the two objectives. 
This requires simpler NN architectures and less training data.

\textbf{\textit{High-level exploration based on Bayesian Optimization: }}
We propose leveraging the capabilities of sequential model-based optimization, such as BO, for efficient exploration of game levels. This approach assumes that the metric of interest is a function of the locations within the level, meaning that coordinates are paired with and mapped to a value of this metric.
For instance, in load tests, agents collect data to identify areas where the metric of interest drops, like GPU or CPU performance or frame rate, for instance due to a high concentration of demanding game entities in a region.
Similarly, for more generic tests that search for crash/freeze states, issues with collision detection, traversability, physics, or any generic test that needs to maximize the map coverage, the metric associated with the coordinate may be a binary variable describing reachability or the frequency of visits. 
This builds either an occupancy map or a heatmap, respectively.

Given that the objective function (metric of interest) is unknown along with its structure, and the goal is to sample it efficiently, BO is well-suited to this search problem.
The unknown function is modelled by the surrogate model using the data the agent collects during the test. 
The BO process then uses this model to compute and optimize the acquisition function, returning the next coordinate to visit. 
The acquisition function balances the exploration-exploitation trade-off by combining the prediction of the surrogate model with the uncertainty estimate.

\textbf{\textit{Grid Map based Surrogate Model: }}In testing time the agent collects data every time frame, hence BO needs to perform its inference with a dataset of millions of samples.
As mentioned earlier, traditional models like GPs or TPEs have limitations and do not scale up for that order of magnitude.
To avoid those constraints, the proposed alternative model condenses all the data to a grid map representation embedded in a 3D array.
The benefit of this representation is that the model size remains constant regardless the growth of the dataset in every time frame, moreover, most of the process can leverage vectorized operations, which are highly efficient.
Thus, regarding the amount of data samples $n$, the complexity of the grid based model is $O(1)$ keeping constant throughout the testing process. 
A prediction of this model is computed simply querying the value of the cell corresponding to the visited coordinate, unlike the case of using a GP whose prediction complexity $O(n^3)$ would make unfeasible to run the system after a couple of minutes of testing.

Additionally, in this representation it is straightforward to incorporate game level information like the NavMesh, explaining what regions are empty or with obstacles.
That is useful for masking the model to discard regions that are not physically meaningful. 
Such integration of the NavMesh into the optimization process for a GP or TPE based surrogate model would be a challenging problem that does not match the minimization techniques used by the BO.

\textbf{\textit{Probabilistic Grid Map: }}
So far it has been introduced the grid map structure of the model and the benefits of it in the map exploration problems.
Nevertheless, BO works under the assumption of the surrogate model being a probabilistic model with the properties of  uncertainty estimation, and smoothness, the latter being necessary for interpolating the collected data for the predictions.
It is proposed to incorporate to the grid map representation the use of Gaussian kernels \eqref{eq:kernel} as often used in GPs and TPEs.
\begin{equation} \label{eq:kernel}
K(x_i, x_j) = \exp\left(-\frac{| x_i - x_j |^2}{2\sigma^2}\right)
\end{equation}
In those models, the kernels are functions that define similarity between two points $x_i$ and $x_j$. 
The kernel bandwidth $\sigma$ determines the sensitivity of that non-linear similarity measure.

The predictions with GPs interpolate the output values in the dataset given the kernel computed similarity of the corresponding input values with the new observation (i.e., the input used for the new prediction).
The uncertainty estimation of GPs works similarly but without considering the output values of the dataset, GPs only measure how the new input vector (in this case the coordinate vector) is in proximity with the input vectors of the dataset. 
Basically a low uncertainty is computed if the input vector is close to other already seen input vectors that are stored in the dataset, because the model has more information about what the output should be, while computing high uncertainty otherwise.

Given that the grid map model has a finite discrete set of possible inputs, this work proposes to pre-compute the prediction and uncertainty estimation for all the map locations.
It determines the similarity and relationship between every possible input and the collected data through a convolution of the grid map and the kernel. 
Such operation returns an interpolation of the collected data, it is a blend of the observed samples weighted by their similarity with respect to the input to be predicted.
The result is the actual surrogate model $f(x,y,z)$, that is a smoothed version of the grid map of the metric of interest, e.g., smoothed heatmap, smoothed performance map. 

Since the uncertainty estimation only requires knowing how the input vector is in proximity of the already observed data, without considering the output values, it can be computed with a similar trick as for the prediction, by convolving the kernel and the occupancy map (binary map indicating the locations already visited), i.e., generating a smoothed occupancy map.
The occupancy map contains the information describing where the agent has been, therefore, the smoothed occupancy map predicts a value close to 1 if a prediction is made in a cell close to already visited points, or close to 0 otherwise.
Such measure opposed to uncertainty can be interpreted as certainty or confidence $c(x,y,z)$, which can be mapped to uncertainty $u(x,y,z)$ with its complement scaled by an amplitude scale parameter defined in the domain of the prediction (equivalent to $\sigma_f$ in a GP).
\begin{equation} \label{eq:uncertainty}
u(x,y,z) = (1-c(x,y,z))\sigma_f
\end{equation}
\textbf{\textit{Acquisition Function:}}
The function to be optimized at every iteration of the BO sets the exploitation-exploration trade off.
This approach balances between exploiting the search of the optimal measure of interest predicted by the surrogate model, and exploring the most unknown areas of the map, determined by the uncertainty estimation.
This function is known as the Lower Confidence Bound (LCB) \eqref{eq:acquisition}, the exploitation-exploration trade off is controlled with the hyperparameter $\sigma_f$ in \eqref{eq:uncertainty}.
\begin{equation} \label{eq:acquisition}
a(x,y,z) = f(x,y,z) - u(x,y,z)
\end{equation}
\textbf{\textit{Low-Level Module Adaptive Exploration: }}
Optimizing the function $a(x,y,z)$ results in the target map coordinates that need to be explored, while the confidence $c(x,y,z)$ determines how much to randomly explore in every possible location. 
The low-level module takes $c(x,y,z)$ as the probability of the agent following the NN policy $\pi(s,x,y,z)$. 
Then, with probability $1-c(x,y,z)$ the agent takes random exploratory actions that help to make the agent perform several ``mistakes", like running into objects or regions that otherwise the NN policy would avoid well, something very useful to find general issues with surrounding game entities. 
The vector $s$ contains the rest of the game states observed by the agent.  
To save time, this adaptive exploration prevents the agent from exploring in already explored areas, and instead traversing them efficiently when they have been well explored already.

\section{Results}
We have been developing and testing the proposed agent testing system exhaustively in first and third person game levels adapted for the specific purpose of this study, both in Unity and Unreal Engine,
while currently we are working to deploy it in a AAA game production.
\begin{figure}
    \centering
    \includegraphics[width=0.60\linewidth]{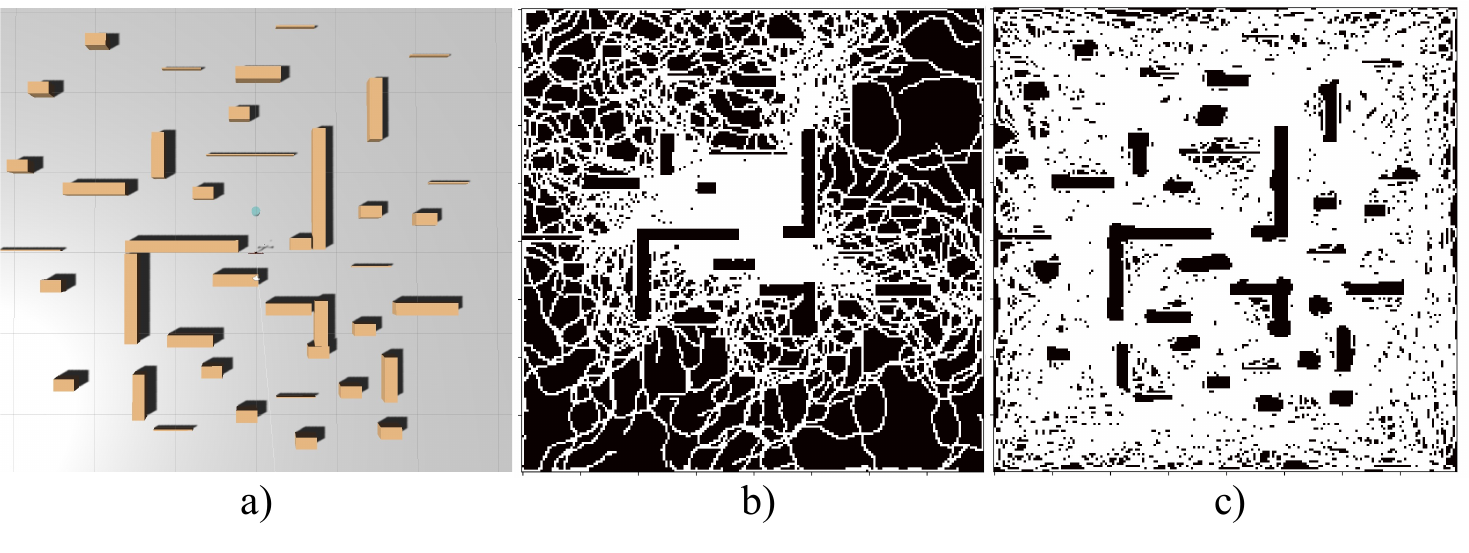}
    \caption{Results of baseline comparison. a) Top view of the small level to cover, and the resulting occupancy maps with b) the baseline, c) the proposed system.   }
    \label{fig:baselinecomparison}
\end{figure}
\begin{figure}
    \centering
    \includegraphics[width=01.0\linewidth]{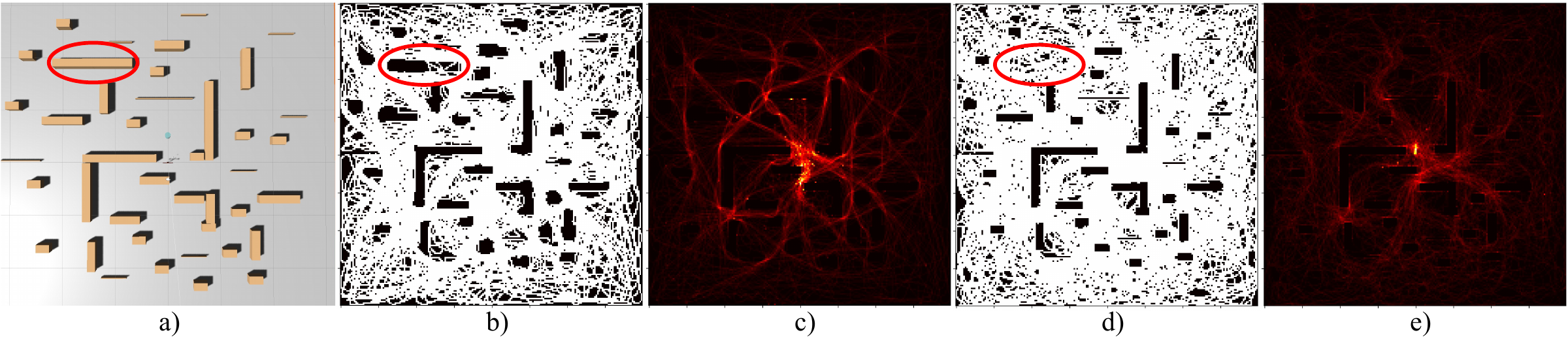}
    \caption{Results of ablating the exploratory actions in the proposed system. a) level highlighting a "ghost wall". Occupancy map and Heatmap in b) and c) from exploring without noisy exploratory actions, and in d) and e) from the full system using random actions determined by the uncertainty of the surrogate model. }
    \label{fig:exploration_ablation}
\end{figure}

The experiments are focused on performing map coverage tests that can trigger any generic bugs.
The baseline for comparison is a system that computes the target exploration point with a uniform random distribution over the level (instead of using BO), and without using any uncertainty measure to take random actions, simply following the NN policy.

In Figure \ref{fig:baselinecomparison}, it is shown a small level used for this comparison, and the occupancy maps obtained by agents testing for the same period of time with the baseline and the proposed system.  
The character is restarted in the center of the scene if it does not reach the target coordinate within a minute.
It is visible how the exploration component of the low-level module helps to cover a wider area while heading to the target, maximising the probabilities of finding bugs, whereas the baseline only does the efficient navigation computed by the low-level module, that is not the most useful in testing time.
Also, it can be seen that the uniform random target selection does not guarantee to cover well the entire level, as there are large empty areas in Figure \ref{fig:baselinecomparison}.b.
The baseline agent spent most of the time around the center of the level as it covered better the central region.

We also simulated a “ghost wall”  that is visible in the level and included in the NavMesh, it is used to evaluate how exploratory actions determined by the model uncertainty are useful to find bugs.
The proposed system is compared to an ablation of itself that does not consider the uncertainty for exploration.
In Figure \ref{fig:exploration_ablation} it is visible that the complete system not only has a slightly better coverage but also went through the “ghost wall” many more times, which is useful for detecting such bug even if that object was considerably smaller.
The heatmaps also show that regardless the agents tend to go through certain corridors to traverse long distances, with the exploratory actions the heatmap is more even.

\begin{table}[]
\caption{Results of ablations of the proposed system. }
\begin{adjustbox}{width=\columnwidth,center}
\begin{tabular}{llllll}
\hline
\textit{Bayesian Optimization}              & TRUE     & TRUE     & TRUE & FALSE    & FALSE                     \\
\textit{Action Exploration}                 & Adaptive & Constant & No   & Adaptive & Constant                  \\ \hline
\textit{Map Coverage } $\uparrow$                          & \textbf{2.52}      & 1.81      & 2.43  & 2.12      & 1.69                       \\ \hline
\textit{Distance to Uniform Distribution 	}$\downarrow$ & \textbf{0.06}     & 0.21     & 0.08 & 0.23     & 0.41 \\ \hline
\end{tabular}
\end{adjustbox}
\label{tab:ablations}
\end{table}
The Table \ref{tab:ablations} contains the main results of ablating the proposed system.
It reports the average results of the percentage of map coverage, and the similarity of the explored distribution with respect to a uniform distribution (full map coverage is a perfect uniform distribution), after testing for an hour with each agent 20 times.
The results are normalized with respect to a system choosing random targets instead of using BO, and not applying exploratory actions.
This video\footnote{\url{https://www.youtube.com/watch?v=3K4jseh3xJw}} shows footage of experiments and operation of this work.

\section{Conclusions}
It is introduced a hierarchical and modular system leveraging the potential of IL/RL agents and BO to explore game levels for testing. 
The proposed grid map based surrogate model makes the BO scalable, it is easy to incorporate additional information like the contained in the NavMesh, and is inexpensive for computing the predictive uncertainty, needed every time frame.
The results showed that agents can attain better map coverage and find bugs more often with such system.




\vspace{12pt}

\bibliographystyle{IEEEtran}
\bibliography{biblio}
\end{document}